\documentclass[sigconf]{acmart}

\usepackage{multirow}  
\usepackage[ruled,vlined]{algorithm2e}
\usepackage{algorithmic}
\usepackage{color}

\newcommand{\mname}{\texttt{BSODA}}

\AtBeginDocument{%
  \providecommand\BibTeX{{%
    \normalfont B\kern-0.5em{\scshape i\kern-0.25em b}\kern-0.8em\TeX}}}

\copyrightyear{2022}
\acmYear{2022}
\setcopyright{rightsretained}
\acmConference[WWW '22] {Proceedings of the ACM Web Conference 2022}{April 25--29, 2022}{Virtual Event, Lyon, France.}
\acmBooktitle{Proceedings of the ACM Web Conference 2022 (WWW '22), April 25--29, 2022, Virtual Event, Lyon, France}

\acmDOI{10.1145/3485447.3512123}
\acmISBN{978-1-4503-9096-5/22/04}

\begin{document}

\title{\mname: A Bipartite Scalable Framework for Online Disease Diagnosis}

\author{Weijie He}
\authornote{Both authors contributed equally to this research.}
\affiliation{%
  \institution{Department of Computer Science and Technology \&
  Institute of Artificial Intelligence \& BNRist, Tsinghua University}
  \city{Beijing}
  \country{China}
}
\email{hwj19@mails.tsinghua.edu.cn}

\author{Xiaohao Mao}
\authornotemark[1]
\affiliation{%
  \institution{Department of Computer Science and Technology \&
  Institute of Artificial Intelligence \& BNRist, Tsinghua University}
  \city{Beijing}
  \country{China}
}
\email{mxh19@mails.tsinghua.edu.cn}

\author{Chao Ma}
\affiliation{%
  \institution{Department of Engineering, University of Cambridge}
  \institution{Microsoft Research Cambridge}
  \city{Cambridge}
  \country{United Kingdom}
}
\email{cm905@cam.ac.uk}

\author{Yu Huang}
\affiliation{%
  \institution{Tencent Jarvis Lab}
  \city{Shenzhen}
  \state{Guangdong}
  \country{China}
}
\email{yorkeyhuang@tencent.com}

\author{Jos{\'e} Miguel Hern{\'a}ndez-Lobato}
\affiliation{%
  \institution{Department of Engineering, University of Cambridge}
  \city{Cambridge}
  \country{United Kingdom}
}
\email{jmh233@cam.ac.uk}

\author{Ting Chen}
\authornote{Corresponding author.}
\affiliation{%
  \institution{Department of Computer Science and Technology \&
  Institute of Artificial Intelligence \& BNRist, Tsinghua University}
  \city{Beijing}
  \country{China}
}
\email{tingchen@tsinghua.edu.cn}


\begin{abstract}
    A growing number of people are seeking healthcare advice online. Usually, they diagnose their medical conditions based on the symptoms they are experiencing, which is also known as self-diagnosis. From the machine learning perspective, online disease diagnosis is a sequential feature (symptom) selection and classification problem. Reinforcement learning (RL) methods are the standard approaches to this type of tasks. Generally, they perform well when the feature space is small, but frequently become inefficient in tasks with a large number of features, such as the self-diagnosis. To address the challenge, we propose a non-RL \underline{B}ipartite \underline{S}calable framework for \underline{O}nline \underline{D}isease di\underline{A}gnosis, called \mname. \mname\ is composed of two cooperative branches that handle symptom-inquiry and disease-diagnosis, respectively. The inquiry branch determines which symptom to collect next by an information-theoretic reward. We employ a Product-of-Experts encoder to significantly improve the handling of partial observations of a large number of features. Besides, we propose several approximation methods to substantially reduce the computational cost of the reward to a level that is acceptable for online services. Additionally, we leverage the diagnosis model to estimate the reward more precisely. For the diagnosis branch, we use a knowledge-guided self-attention model to perform predictions. In particular, \mname\ determines when to stop inquiry and output predictions using both the inquiry and diagnosis models. We demonstrate that \mname\ outperforms the state-of-the-art methods on several public datasets. Moreover, we propose a novel evaluation method to test the transferability of symptom checking methods from synthetic to real-world tasks. Compared to existing RL baselines, \mname\ is more effectively scalable to large search spaces.

\end{abstract}



\maketitle

\section{Introduction}
The explosive development of the mobile web has increased the feasibility and potential of access to online disease diagnosis. A growing number of adults are now attempting to gain a better understanding of their health conditions through online services before visiting a doctor. According to a national survey conducted in the United States \cite{fox2013health}, 35\% of American adults regularly use the internet to self-diagnose. Numerous individuals have long relied on search engines to access online health information \cite{fox2013health}. However, current search technology is often ineffective for self-diagnosis \cite{zuccon2015diagnose}. The fact that the majority of users lack adequate medical knowledge \cite{keselman2008consumer} has an effect on the quality of health queries \cite{hu2010search, lopes2013query, puspitasari2017impacts}.

In recent years, many symptom checking tools have been developed as an alternative to online self-diagnosis \cite{semigran2015evaluation}, such as the well-known symptom checkers from Mayo Clinic \cite{mayo} and WebMD \cite{webmd}. A symptom checker simulates the disease diagnosis process which \citet{ledley1959reasoning} summarized into three steps: 1) A patient presents a self-report of initial symptoms; 2) A doctor inquires the patient with a series of questions regarding additional symptoms, history, and other related information; and 3) The doctor makes a final diagnosis. When it comes to online self-diagnosis, the questions would concern patients' symptoms.
Compared to search engines, symptom checkers have the advantage of not requiring users to formulate appropriate health queries, which they are incapable of \cite{zeng2004positive, zuccon2015diagnose}. Besides, symptom checkers are more accurate in disease diagnosis and require less effort and time.

Although self-diagnosis through a symptom checker is a substantial advance over patient-directed internet searches, it should be highlighted that symptom checkers are not a substitute for medical investigation by a doctor. Physical examinations and diagnostic tests such as X-rays and laboratory tests, which are critical in the diagnosis of many diseases, could not be performed by symptom checkers. However, there is indeed a genuine need in the field of healthcare to develop a symptom checker that is user-friendly, efficient, and accurate \cite{cross2021search}. 

From the healthcare perspective, the primary goal of symptom checking is diagnosis accuracy. A symptom checker acquires sequentially a number of symptom values in order to make an accurate diagnosis. Naturally, more information about possible symptoms could result in a more accurate diagnosis; for example, most existing questionnaire systems acquire a large number of symptom values from patients through exhaustive inquiries \cite{shim2018joint,lewenberg2017knowing}. Unfortunately, both patients and experts find it inefficient, time-consuming, and unbearable. The actual goal of symptom checking is to minimize the number of inquiries while achieving a high diagnostic accuracy.

From the machine learning perspective, symptom checking can be viewed as a cost-sensitive sequential feature selection and classification task. Acquiring a symptom’s value can be viewed as selecting a feature which comes with a cost.
Reinforcement learning (RL) methods have shown good performance in this type of tasks with a small set of features \cite{tang2016inquire, kachuee2018opportunistic, kao2018context, peng2018refuel, xia2020generative, liao2020task, lin2020towards, liu2021dialogue}. However, they often suffer in tasks with large feature spaces, because they struggle to train a model. EDDI \cite{ma2019eddi}, a framework for instance-based active feature acquisition, proposes an approach to choose the next feature by maximizing a defined information reward over all features. It uses a Partial Variational AutoEncoder (VAE) \cite{kingma2013auto} to handle data with missing values. Such a model can deal with the situation in which an agent only knows partial information of patients when inquiring. Although promising, EDDI is limited by its high computational cost dealing with a large number of symptoms.

To address the aforementioned issues, we propose \mname, a \underline{B}ipartite \underline{S}calable framework for \underline{O}nline \underline{D}isease di\underline{A}gnosis. Different from prior methods, \mname\ is a novel non-RL approach to automatic symptom-inquiry and disease-diagnosis. RL methods that perform both symptom-inquiry and disease-diagnosis simultaneously must deal with a large heterogeneous action space that poses a challenge to learning with efficiency and effectiveness \cite{liu2021dialogue}. Yet \mname\ divides the task into two cooperative branches: inquiry and diagnosis. The inquiry branch uses a generative model to produce an inquiry, while the diagnosis branch uses a supervised classification model to predict the disease. \mname\ leverages the characteristics of these two models so that they cooperate in decision-making. 

Specifically, the inquiry branch determines which symptom to collect next based on the information-theoretic reward function that was used in EDDI \cite{ma2019eddi}. Instead of using the Partial VAE, \mname\ employs a Product-of-Experts encoder to better handle partial observations of symptoms, which has shown good performance with missing multimodal data \cite{wu2018multimodal}. Besides, to estimate the reward more precisely, a novel two-step sampling strategy is developed to approximate the joint posterior by taking advantage of the predictive probability distributions from the diagnosis model. Special strategies are designed to accelerate the computation of \mname.

For the diagnosis branch, \mname\ returns predictive disease distributions by a knowledge-guided self-attention model that has been shown to be efficient at learning the graphical structure of electronic health record \cite{choi2020learning}. \mname\ leverages this model to embed the relationships between diseases and symptoms to incorporate prior knowledge. In particular, \mname\ samples unobserved features from the generative model to model the uncertainty caused by partial observations of features, which assists the diagnosis model in making predictions of the diagnosis results and determining when to statistically terminate the inquiry process.

Unlike most previous RL baselines that have been proved to be effective exclusively on synthetic or real-world datasets, \mname\ performs well on both types of evaluations. Synthetic datasets were developed for model evaluation because obtaining a large amount of real-world medical data for training is difficult. However, there is a significant gap between existing synthetic and real-world datasets. Good performance on synthetic datasets is not a guarantee of transferability to the real world. Hence we propose a novel approach to evaluating the model transferability using both synthetic and real-world tasks. Synthetic data consisting of thousands of features are used in model training to assess a model’s ability to cope with large search spaces, while real-world data is only used for testing to assess the transferability of the trained model.

In summary, our main contributions are highlighted as: 1) We propose \mname, a non-RL bipartite framework for online disease diagnosis. It is scalable to large feature spaces at a low computational cost, where RL methods typically struggle. It demonstrates a high degree of generalizability across a variety of settings; 2) Technically, we design the models and workflow for \mname\ in order to facilitate collaboration between the symptom-inquiry and disease-diagnosis branches. We also develop several speedup techniques to make \mname\ practical for online services; and 3) We propose a novel evaluation method and new datasets to assess the performance of symptom checking methods on large search spaces and transferability to the real-world tasks.

\section{Related Work}

There are a significant amount of works that deal with self-diagnosis. One large family of Bayesian inference and tree-based methods \cite{ kononenko1993inductive,kononenko2001machine,xu2013cost,kohavi1996scaling} uses entropy functions to select symptoms based on the theory of information gain. For instance, \citet{ nan2015feature,nan2016pruning} and \citet{ zakim2008underutilization} proposed to address the cost of feature acquisition by using decision tree and random forest methods. \citet{hayashi1991neural} attempted to extract rule-based representations from medical data and human knowledge to perform diagnosis. \citet{early2016test} and \citet{kachuee2018dynamic} suggested performing a sensitivity analysis on trained predictors to measure the importance of each feature in a given context. \citet{wang2021online} developed a hierarchical lookup framework based on the symptom embedding where graph representation learning is tailored towards disease diagnosis. Due to the intractable nature of global maximization of information gain or global sensitivity, these methods often employ greedy or approximation algorithms that result in low accuracy. 

Recently, \citet{janisch2019classification, janisch2020classification} showed that reinforcement learning (RL) methods outperformed tree-based methods in the task of sequential feature acquisition with a specific cost function. In particular, \citet{tang2016inquire} first formulated the inquiry and diagnosis process as a Markov decision process and then used RL to perform symptom checking in a simulated environment. \citet{kao2018context} and \citet{peng2018refuel} demonstrated that competitive results could be obtained even in medium search spaces. \citet{kachuee2018opportunistic} proposed a method based on deep Q-learning with variations of model uncertainty as the reward function. \citet{xia2020generative} implemented an RL agent for symptom checking using Generative Adversarial Network (GAN) and policy gradient method, which performed well on two public datasets containing instances of four and five diseases, respectively. However, such a small number of diseases is impractical in most real-world situations. \citet{lin2020towards} proposed to combine an inquiry branch of a Q network with a diagnosis branch that involves a generative sampler. It was only effective on the two small datasets mentioned previously. \citet{liao2020task} proposed a hierarchical RL framework with a master for worker appointment, several workers for symptom inquiry, and a separate worker for screening diseases. \citet{liu2021dialogue} improved it by using a pretraining strategy to overcome difficulties in convergence. However, their maximum manageable number of diseases is 90, and the model did not perform well in current datasets.  

\section{Methodology}

In this section, we first present the technical background of \mname\ and then introduce the inquiry branch, including the PoE encoder, the new sampling scheme, and the acceleration methods for reward estimation. Next, we develop a knowledge-guided self-attention model in the diagnosis branch. At the end, we present the diagnostic method that incorporates these two branches and the termination criterion of the inquiry and diagnosis process. Figure~\ref{fig:flowchart} shows the \mname\ process during a single round of inquiry and diagnosis. 

\subsection{Background}
\label{sec:bg}
\subsubsection{Problem Formulation}
In this paper, we formulate symptom checking as a sequential feature selection and classification problem. Let $S$ denote the set of all possible symptoms, and $D$ denote the set of possible diseases. Suppose $x_s$ denotes the presence or absence of symptom $s \in S$: $x_s = 1$ if a patient actually suffers from the symptom $s$, i.e., \emph{positive symptom}, and $x_s = 0$ otherwise. Similarly, $\mathbf{x}_D$ denotes the one-hot categorical vector of which each dimension indicates whether a disease is present (= 1) or not (= 0). Then, we are interested in predicting the target variables $\mathbf{x}_D$  given the corresponding observed features $\mathbf{x}_O$, where $O\subset S$ is the set of symptoms that are currently observed, and $U = S\setminus O$ denotes the set of the unobserved ones. More specifically, we choose which symptom $x_s \in \mathbf{x}_{U}$ to inquire next so that our belief regarding $\mathbf{x}_D$ can be optimally improved. Notations are summarized in Table~\ref{tab:not}.

\begin{table}
  \caption{Notations.}
  \label{tab:not}
  \begin{tabular}{cc}
    \toprule
    Notation & Description\\
    \midrule
    $S/D$ & Set of symptoms/diseases.\\
    $S_c$ & Set of candidate symptoms by filtering.\\
    $x_s$ & Presence or absence of symptom $s$ ($s \in S$).\\
    $\mathbf{x}_D$ & Presences or absences of diseases.\\
    $\mathbf{x}_O/\mathbf{x}_U$ & Observed/Unobserved symptoms.\\
    $\mathbf{x}$ & Complete data ($\mathbf{x}=\mathbf{x}_S\cup\mathbf{x}_D=\mathbf{x}_O\cup\mathbf{x}_U\cup\mathbf{x}_D$).\\
    $N_T$ & Maximum number of inquiries.\\
    $N_C$ & Number of candidate symptoms.\\
    $N_M$ & Number of Monte Carlo samples.\\
    $\Phi/\theta$ & Parameters of the VAE encoder/decoder.\\ 
    $p_D(\cdot)$ & Predictive distribution from the diagnosis model.\\
    $\mathbf{M}$ & Mask matrix.\\
    $\mathbf{P}$ & Conditional probability matrix.\\
    $\mathbf{A}$ & Attention matrix.\\
    $\mathbf{e}/\mathbf{E}$ & Embedding vector(s) of feature(s).\\
  \bottomrule
\end{tabular}
\end{table}

\begin{figure*}
	\centering
	\includegraphics[width=\linewidth]{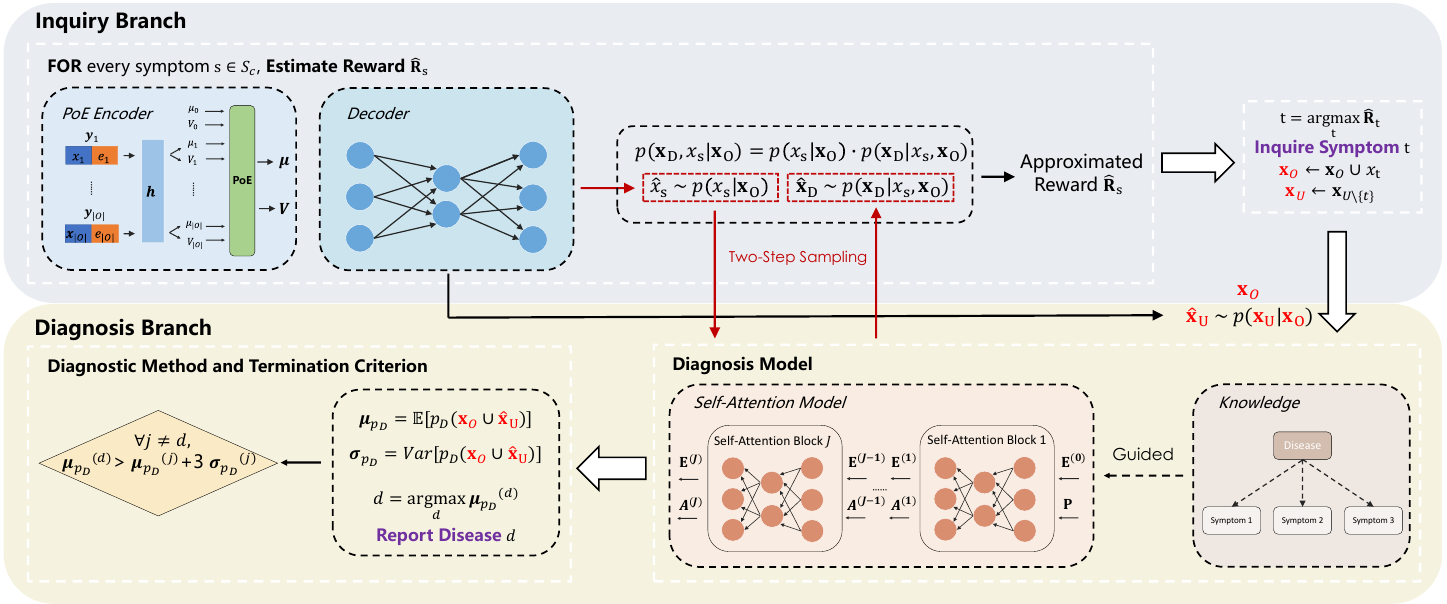}
	\caption{Overall flowchart of \mname\ with one round of inquiry and diagnosis.}
	\Description{At first, observed features $\mathbf{x}_O$ are fed into the inquiry model. For every possible symptom $s$ in $S_c$ that is built by the filtering strategy according to prior knowledge, the inquiry branch estimates the reward $\hat{R}_s$, which applies a two-step sampling strategy and approximation methods. Next, the inquiry branch inquires the symptom $t$ and updates $\mathbf{x}_O$ and $\mathbf{x}_U$, where $t=\text{argmax}_t\hat{R}_t$. After that, $\mathbf{x}_O$ and $\hat{\mathbf{x}}_U\sim p(\mathbf{x}_U|\mathbf{x}_O)$ from the inquiry branch are fed into the diagnosis model to make predictions. Finally, the termination criterion determines whether to continue inquiring symptoms.}
	\label{fig:flowchart}
\end{figure*}

\subsubsection{EDDI}
A variational autoencoder (VAE) \cite{kingma2013auto} defines a generative model of the form $p(\mathbf{x}, \mathbf{z})=\prod_i p_\theta(x_i|\mathbf{z})p(\mathbf{z})$ in which the data $\mathbf{x}$ is generated from latent variables $\mathbf{z}$, $p(\mathbf{z})$ is a prior, e.g., spherical Gaussian, and $p_\theta(\mathbf{x}|\mathbf{z})$ is presented as a neural network decoder with parameters $\theta$ to specify a simple likelihood, e.g., Bernoulli. A VAE uses another neural network with parameters $\Phi$ as an encoder to produce a variational approximation of the posterior, that is, $q_\Phi(\mathbf{z}|\mathbf{x})$. A VAE is trained by maximizing an evidence lower bound (ELBO):
\begin{equation}
    \mathbb{E}_{q_\Phi(\mathbf{z}|\mathbf{x})}[\log p_\theta(\mathbf{x}|\mathbf{z})]-\beta\cdot D_\textrm{KL}[q_\Phi(\mathbf{z}|\mathbf{x})\,\|\, p(\mathbf{z})],
\end{equation}
where $\beta$ is the weight to balance the two terms in the expression and $D_{\textrm{KL}}$ is the Kullback-Leibler (KL) divergence between two distributions. The ELBO is usually optimized using stochastic gradient descent and the reparameterization trick \cite{kingma2013auto}. EDDI \cite{ma2019eddi} is a recently proposed VAE-based framework for the feature acquisition problem. EDDI chooses the next feature $x_s$ to acquire by maximizing over the information reward
\begin{equation} \label{eq:IR}
R_s = \mathbb{E}_{x_s \sim p(x_s|\mathbf{x}_O)}
	D_{\textrm{KL}}\left[p(\mathbf{x}_D | x_s,\mathbf{x}_O) \,\|\, p(\mathbf{x}_D | \mathbf{x}_O)
\right].
\end{equation}
The VAE is used to express the conditional probabilities $p(x_s|\mathbf{x}_O)$, $p(\mathbf{x}_D|\mathbf{x}_s,x_O)$, and $p(\mathbf{x}_D|\mathbf{x}_O)$. Unfortunately, estimating the values of these quantities in Eqn.~\ref{eq:IR} is prohibitively expensive. To address this problem, \citet{ma2019eddi} demonstrate how Eqn.~\ref{eq:IR} can be efficiently approximated using VAE encoding distributions:
\begin{equation}
    \begin{aligned}
    \label{eq:r}
    \hat{R}_s=\mathbb{E}_{x_s\sim p_\theta(x_s|\mathbf{x}_O)}&D_\textrm{KL}[q_\Phi(\mathbf{z}|x_s,\mathbf{x}_O)\,\|\,q_\Phi(\mathbf{z}|\mathbf{x}_O)]-\\
    \mathbb{E}_{\mathbf{x}_D ,x_s\sim p_\theta(\mathbf{x}_D ,x_s|\mathbf{x}_O)}D_\textrm{KL}&[q_\Phi(\mathbf{z}|\mathbf{x}_D ,x_s,\mathbf{x}_O)\,\|\,q_\Phi(\mathbf{z}|\mathbf{x}_D ,\mathbf{x}_O)].
    \end{aligned}
\end{equation}
The first expectation of Eqn. \ref{eq:r} quantifies how much information that $x_s$ provides about $\mathbf{z}$, whereas the second expectation quantifies how much information that $x_s$ provides about $\mathbf{z}$ with an addition of $\mathbf{x}_D$. A feature $x_s$ will be penalized by the second term if it is informative about $\mathbf{z}$ but not about $\mathbf{x}_D$. All quantities in the KL divergences above can be computed analytically using Gaussian approximations. The expectations in Eqn. \ref{eq:r} can be approximated by a Monte Carlo process, averaging across samples $\hat{\mathbf{x}}_D, \hat{x}_s\sim p(\mathbf{x}_D, x_s|\mathbf{x}_O)$ that can be shared between the two expectations.

\subsection{Inquiry Branch}

We adopt EDDI \cite{ma2019eddi} as the backbone of the inquiry branch. The encoder of the VAE is used to handle partial observations of features during active feature selection. Let $h(\cdot)$ denote a neural network, $g(\cdot)$ denote the summation or max-pooling operation, $\mathbf{e}_i$ denote the embedding vector of the $i$-th feature, and $\mathbf{y}_i$ denote the input carrying information of the $i$-th observed feature, which is constructed by concatenation: $\mathbf{y}_i=[x_i, \mathbf{e}_i]$. EDDI proposed a permutation invariant set function as an encoder, given by $\mathbf{c}(\mathbf{x}_O):=g(h(\mathbf{y}_1),h(\mathbf{y}_2),...,h(\mathbf{y}_{|O|}))$, where $|O|$ denotes the number of observed features. Finally, the code $\mathbf{c}(\mathbf{x}_O)$ is fed into a neural network and transformed into the statistics of a multivariate Gaussian distribution to approximate $p(\mathbf{z}|\mathbf{x}_O)$.

\subsubsection{Product-of-Experts (PoE) Encoder} A simple $g(\cdot)$ cannot handle a large number of features. In \mname, we propose to use a PoE encoder \cite{wu2018multimodal} to approximate the joint posterior of the latent variables. It assumes conditional independence among features. The approximated posterior, including a prior Gaussian expert $p(\mathbf{z})$ with mean vector $\mu_0$ and variance vector $V_0$, is given by
\begin{equation}
\label{eq:mvae}
    q_\Phi(\mathbf{z}|\mathbf{x}_O)\propto p(\mathbf{z})\prod q_\Phi(\mathbf{z}|x_i),
\end{equation}
where $q_\Phi(\mathbf{z}|x_i)$ is an inference network denoting the expert associated with the $i$-th observed feature $x_i$. Figure~\ref{fig:flowchart} illustrates the architecture of the PoE encoder. $\mathbf{e}_i$ is pretrained by the diagnosis model and fixed during training (please refer to Sec.~\ref{sec:diag_model}). Then we use a Multiple Layer Perceptron (MLP) as $h(\cdot)$ to map the input $\mathbf{y}_i$ to a Gaussian distribution in latent space with mean vector $\mu_i$ and variance vector $V_i$. $h(\cdot)$ shares parameters among all features. Because a product of Gaussian experts is itself Gaussian \cite{cao2014generalized}, we can quickly compute variance $V=(V_0^{-1}+\sum V_i^{-1})^{-1}$ and mean $\mu=(\mu_0 V_0^{-1}+\sum\mu_i V_i^{-1})V$ for $q_\Phi(\mathbf{z}|\mathbf{x}_O)$ in Eqn.~\ref{eq:mvae}. The decoder, $p_\theta(\mathbf{x}|\mathbf{z})$, is given by a product of Bernoulli distributions, the probabilities of which are specified by an MLP that accepts $\mathbf{z}$ as input.

If we train the PoE encoder on all available data for a dataset with no missing features, its performance would suffer when the input contains missing entries, since the PoE encoder has never encountered such a case during training. To address this, we drop a random fraction of the fully observed features $\mathbf{x}$, which includes both $\mathbf{x}_S$ and $\mathbf{x}_D$, for each data point during training.

\subsubsection{Two-Step Sampling Strategy} 
\label{sec:two-step}
To estimate the information reward, we approximate expectations by Monte Carlo sampling, averaging across the samples $\hat{\mathbf{x}}_{D}, \hat{x}_s\sim p(\mathbf{x}_{D}, x_s|\mathbf{x}_O)$. This can be accomplished by first sampling $\hat{\mathbf{z}}\sim q_\Phi(\mathbf{z}|\mathbf{x}_O)$, followed by $\hat{\mathbf{x}}_{D}, \hat{x}_s\sim p_\theta(\mathbf{x}_{D}, x_s|\hat{\mathbf{z}})$. However, we propose a more accurate method for \mname, in which $\hat{x}_s$ and $\hat{\mathbf{x}}_{D}$ are sampled in two steps. Since
\begin{equation}
    \label{eq:two-step}
    p(\mathbf{x}_D, x_s|\mathbf{x}_O)=p(\mathbf{x}_D|x_s, \mathbf{x}_O)\cdot p(x_s|\mathbf{x}_O),
\end{equation}
we propose to sample $\hat{x}_s$ from the VAE by sampling $\hat{\mathbf{z}}\sim q_\Phi(\mathbf{z}|\mathbf{x}_O)$, and then $\hat{x}_s\sim p_\theta(x_s|\hat{\mathbf{z}})$. Next we sample from $\hat{\mathbf{x}}_D\sim p_D(\hat{x}_s,\mathbf{x}_O)$ produced by a diagnosis model. In this way, we use two networks, the VAE and a diagnosis model, to approximate the joint posterior. By combining the generative and classification models, we can improve the performance of reward estimation.

\subsection{Speedup for Reward Estimation}
\label{sec:speedup}
The computational cost of EDDI \cite{ma2019eddi} to estimate the reward is $O(N_T\cdot N_C\cdot N_M)$, where $N_T$ denotes the maximum number of inquiries, $N_C$ denotes the total number of candidate symptoms, and $N_M$ denotes the number of samples required in the Monte Carlo process. Because $N_C$ can account for thousands of symptoms, it would be too expensive to support an online disease diagnosis service. This section shows how we may significantly reduce the cost by taking advantage of feature sparsity and predictive probability distributions from the diagnosis model.

\subsubsection{Filtering Candidate Symptoms by Prior Knowledge} 
\label{sec:fis}
In each round, the model chooses a feature $x_s$ from $\mathbf{x}_U$ in order to maximize the reward. Under the setting of symptom-inquiry, we can filter out irrelevant symptoms to reduce the number of candidates that need to be queried. 

For every symptom $s$, we calculate the set $S_s$ of additional symptoms that may co-occur with symptom $s$ with high probability in one patient, based on prior knowledge or training data statistics. More specifically, we assign a symptom $j \in S_s$ if $P_{data}(x_j=1|x_s=1)$ is larger than a pre-defined threshold, which we can simply set to 0 by default. At the beginning of the inquiry process when symptom $i$ is present, the set of candidate symptoms, $S_c$, is initialized to $S_i$. Every time the framework selects a new positive symptom $j$, we update the set of candidate symptoms to be the intersection of the current set and $S_j$. Thus we reduce the computational cost $O(N_T\cdot N_C\cdot N_M)$ by lowering $N_C$.

\subsubsection{Approximation of Reward}
\label{sec:app}
The reward $\hat{R}_s$ in Eqn. \ref{eq:r} is estimated by averaging over $N_M$ Monte Carlo samples $\hat{\mathbf{x}}_D, \hat{x}_s\sim p(\mathbf{x}_D, x_s|\mathbf{x}_O)$. It should be noted that $x_s$ is binary and $\mathbf{x}_D$ is discrete among $|D|$ diseases in symptom checking tasks. Indeed we just need to compute rewards of $2\cdot |D|$ distinct combinations of $x_s$ and $\mathbf{x}_D$, each with a weight $p(\mathbf{x}_D, x_s|\mathbf{x}_O)$ that is solved using the two-step sampling strategy described in Sec.~\ref{sec:two-step}. Then the computational cost is $O(N_T\cdot N_C\cdot (2 |D|))$. In the following, we demonstrate how to speed up it by reducing $2 |D|$. 

Firstly, consider that the number of symptoms experienced by a patient with a specific disease is generally smaller than the number of possible symptoms for the disease, which is much smaller than the total number of symptoms $|S|$. With such a sparse feature space, the inquiry process should be prioritized to acquire positive symptoms \cite{peng2018refuel}. Motivated by this, we encourage \mname\ to focus on positive symptoms by discarding combinations of $x_s$ and $\mathbf{x}_D$ when $\hat{x}_s=0$, i.e., setting their rewards to 0. Hence the number of combinations of $x_s$ and $\mathbf{x}_D$ to be computed is reduced to $|D|$. 

Secondly, when $|D|$ is large, practically almost all predictive distributions $p_D(\cdot)$ are long-tailed. Hence we rank all possible disease values for $\mathbf{x}_D$ by its probability in $p_D(\hat{x}_s,\mathbf{x}_O)$, using the two-step sampling strategy in Sec.~\ref{sec:two-step}, and we discard those that fall below the 90 percent of the top probability or are less than $1 / |D|$. Then the average number of combinations to be computed can be further reduced to less than 10 experimentally.

\subsection{Diagnosis Branch}
\label{sec:diag}

It should be noted that we may wish to perform diagnosis in any round of the symptom checking process. 
Recently, an increasing number of RL methods \cite{lin2020towards, liao2020task, liu2021dialogue} have used a supervised learning model to handle the diagnosis. In \mname, we denote $p_D(\cdot)$ to be the disease predictive distributions returned by the diagnosis model.

\subsubsection{Knowledge-Guided Self-Attention Model}
\label{sec:diag_model}

\mname\ uses a self-attention mechanism guided by prior knowledge to embed relationships between all features, consisting of both diseases and symptoms. We use two matrices, $\mathbf{M}$ and $\mathbf{P}$, to represent prior knowledge, which are of the same size $(|S|+|D|)\times(|S|+|D|)$. $\mathbf{M}$ is a mask matrix of which each entry $\mathbf{M}_{ij}=0$ denotes that the relationship between feature $i$ and feature $j$ is considered, and negative infinity otherwise. Specifically, in $\mathbf{M}$, only disease-disease relationships are not considered. $\mathbf{P}$ is a matrix of which each entry $\mathbf{P}_{ij}$ denotes the conditional probability $P(j|i)$, which is calculated by co-occurrence between $i$ and $j$ according to prior knowledge or training data statistics, and normalized such that each row sums to 1, i.e., $P(\cdot|i)=1$. Again, only $P(\text{symptom}|\text{disease})$ and $P(\text{symptom}|\text{symptom})$ are calculated, and all other probabilities are set to zeros. 

The self-attention model consists of several stacked self-attention blocks. Each block accepts the outputs from the previous block, including embeddings of all features and an attention matrix representing the weights of the relationships between features, and calculates new embeddings and attention for the next block. The output of an attention function can be described as a weighted sum of the $\mathbf{V}$alues, where the weights are computed by the $\mathbf{Q}$uery with the corresponding $\mathbf{K}$ey \cite{vaswani2017attention}. For self-attention, $\mathbf{Q}$, $\mathbf{K}$ and $\mathbf{V}$ are derived from the same source. Defining $\mathbf{A}^{(j)}$ as the attention matrix and $\mathbf{E}^{(j)}$ as the embedding vectors calculated by the $j$-th block ($j\ge1$), the self-attention block is given by:
\begin{equation}
    \label{eq:cla}
    \begin{aligned}
    \mathbf{A}^{(j)}&=\text{Softmax}(\frac{\mathbf{Q}^{(j)}\mathbf{K}^{(j)\top}}{\sqrt{c}}+\mathbf{M}),\\
    &\mathbf{E}^{(j)}=\text{MLP}^{(j)}(\mathbf{A}^{(j)}\mathbf{V}^{(j)}),\\
    \mathbf{Q}^{(j)}=\mathbf{E}^{(j-1)}\mathbf{W}_Q^{(j)}&,\  \mathbf{K}^{(j)}=\mathbf{E}^{(j-1)}\mathbf{W}_K^{(j)},\ \mathbf{V}^{(j)}=\mathbf{E}^{(j-1)}\mathbf{W}_V^{(j)},\\
    \end{aligned}
\end{equation}
where $c$ is the column size of $\mathbf{W}_K$. All $\mathbf{W}$s are trainable parameters. $\mathbf{E}^{(0)}$ is constructed by concatenation: if $i\in S$, $\mathbf{E}_i^{(0)}=[x_s,\mathbf{e}_i]$; if $i\in D$, $\mathbf{E}_i^{(0)}=[0,\mathbf{e}_i]$. When calculating the attention matrix $\mathbf{A}$, the attention weights of the relationships that are not considered are fixed to 0 by the negative infinities in $\mathbf{M}$. To incorporate prior knowledge $\mathbf{P}$, we initialize $\mathbf{A}^{(0)}$ by $\mathbf{P}$ and sequentially penalize the attention of the $j$-th block if it deviates too much from the previous block by a regularized training loss term, $\sum_j D_\textrm{KL}(\mathbf{A}^{(j-1)}||\mathbf{A}^{(j)})$. Finally, the predictive distribution is obtained from the last embedding vectors of $\mathbf{x}_{D}$: $p_D(\cdot)=\text{Softmax}(\text{MLP}^D(\mathbf{E}_D^{(J)}))$.
We train the model on complete data $\mathbf{x}$, which accepts $\mathbf{x}_S$ as input and predicts $\mathbf{x}_D$. The learned embeddings $\mathbf{E}$ are used to initialize the VAE model. 

\subsubsection{Diagnostic Method and Termination Criterion}
\label{sec:stop}
The uncertainty caused by partial observations of features will make it difficult to perform an accurate diagnosis. Hence, to model the uncertainty, we impute the missing variables $\mathbf{x}_{U}$ in the input by drawing $N_M$ samples from $\hat{\mathbf{x}}_{U}\sim p(\mathbf{x}_{U}|\mathbf{x}_O)$, for which the sampling process is the same as $\hat{x}_s\sim p(x_s|\mathbf{x}_O)$ according to the two-step sampling strategy in Sec.~\ref{sec:two-step}. These $N_M$ samples $\{ (\hat{\mathbf{x}}_{U},\mathbf{x}_O ) \}_{m=1}^{N_M}$ are then fed into the diagnosis model, yielding a set of predictive distributions on $\mathbf{x}_D$, denoted by $\{(p^{(1)},...,p^{(|D|)})^{m}\}_{m=1}^{N_M}$. Then, we calculate the expectations $\mu_{p_D} = \mathbb{E} [p^{(1)},...,p^{(|D|)}]  $ and standard deviations $\sigma_{p_D} = \sqrt{\mathrm{Var} [p^{(1)},...,p^{(|D|)}]}$. \mname\ would stop inquiring to report the chosen disease if the probability of this disease is so high that inquiring more symptoms would not overturn the diagnostic result \cite{lin2020towards}. That is, the inquiry process would stop when the probability of the chosen disease is beyond the upper bound of the $6\sigma$ interval \cite{parzen1960modern,2001What} of the probabilities of other diseases. When a single disease $d$, where $d=\text{argmax}_d\ \mu_{p_D}^{(d)}$, is chosen, the termination criterion can be formulated as:
\begin{equation}
    \label{eq:stop}
    \forall j\neq d, \mu_{p_D}^{(d)}>\mu_{p_D}^{(j)}+3\sigma_{p_D}^{(j)}.
\end{equation}

\section{Experiments}
\label{sec:exp}

\begin{table}\scriptsize
\caption{Essential characteristics of experimental datasets.}  
\label{tab:ds}

\begin{tabular}{cccccc}  
\toprule  
Dataset & Type & Data Size & \#Diseases & \#Symptoms & Usage \\
\midrule
SymCAT & Synthetic & - & 801 & 474 & Train \& Test\\
\midrule
MuZhi & Real-World & 710 & 4 & 66 & Train \& Test\\
Dxy & Real-World & 527 & 5 & 41 & Train \& Test\\
\midrule
HPO-Synthetic & Synthetic & - & 11,441 & 13,032 & Train \& Test\\
HPO-HMS & Real-World & 83 & 37 & 747 & Test\\
HPO-MME & Real-World & 43 & 18 & 559 & Test\\
\bottomrule  
\end{tabular}  

\end{table}

Table~\ref{tab:ds} summarizes the essential characteristics of the experimental datasets. \mname\ is evaluated on three categories of experimental datasets including a synthetic dataset, two real-world common disease datasets, and three rare disease datasets including one synthetic dataset derived from a knowledge base and two real-world datasets. They are all publicly accessible and do not contain personally identifiable information. We use synthetic datasets to generate several synthetic tasks with different numbers of possible diseases. 

\subsection{Existing Datasets}

\subsubsection{SymCAT}

SymCAT is a symptom-disease database \cite{symcat}. For each disease, there is information about its symptoms with their marginal probabilities. It is filtered using the occurrence rates in Centers for Disease Control and Prevention (CDC) database. Following \citet{peng2018refuel}, we use SymCAT to build a synthetic dataset. We first sample a disease and its related symptoms from all diseases uniformly at random. Then, we perform a Bernoulli trial on each extracted symptom based on its associated probability to form a symptom set. For instance, if we sample a disease "abscess of nose", we would obtain its associated symptoms: "cough" and "fever", which occur with probabilities of 73\% and 62\%, respectively. We then generate one data instance by sampling Bernoulli random variables according to these probabilities. We sample $10^6$, $10^5$, and $10^4$ records for training, validation, and testing, respectively.

\subsubsection{Two Real-World Datasets}

\citet{wei2018task} constructed the MuZhi Medical Dialogue dataset, which was collected from the dialogue data from the pediatric department of a Chinese online healthcare website (\url{https://muzhi.baidu.com/}). The MuZhi dataset includes 66 symptoms for 4 diseases: children's bronchitis, children's functional dyspepsia, infantile diarrhea infection, and upper respiratory infection. 

The Dxy Medical Dialogue dataset \cite{xu2019end} contains data from another popular Chinese online healthcare website (\url{https://dxy.com/}), where users often seek professional medical advice from specialists. The Dxy dataset includes 41 symptoms for 5 diseases: allergic rhinitis, upper respiratory infection, pneumonia, children hand-foot-mouth disease, and pediatric diarrhea. These two datasets have been structured in the form of disease-symptoms pairs and split into a training set and a testing set. 

\subsection{Proposed HPO-Based Rare Disease Datasets}

Human Phenotype Ontology (HPO) \cite{kohler2017human} provides a standardized vocabulary of phenotypic abnormalities associated with human disease. The initial domain application of the HPO was on rare disorders. There are 123,724 annotations of HPO terms to rare diseases and 132,620 to common diseases. These were compiled from multiple sources, including the medical literature, Orphanet (\url{https://www.orpha.net/}), OMIM (\url{https://omim.org/}) and DECIPHER (\url{https://decipher.sanger.ac.uk/}). We chose diseases and related symptoms from the public rare disease dictionary, along with their marginal probabilities, to build new datasets. 

The HMS dataset was collected by \citet{knitza2019german}, including 93 rare disease cases from original real-world health records. The Matchmaker Exchange API \cite{buske2015matchmaker} provides a shared language that databases can use to query each other to find similar patients. The API is built around a standardized patient profile including both phenotype and genotype information. The MME dataset is the standardized rare disease test set for the API, containing 50 de-identified patients selected from publications. We chose 83 and 43 cases from the HMS and MME datasets, respectively, to form new HPO-HMS and HPO-MME datasets. Each of them have at least one symptom matching a particular HPO term.

We propose a novel HPO-based evaluation approach using both synthetic and real-world datasets. The HPO database is used to create a synthetic dataset as the simulation process of SymCAT. The real-world datasets, HPO-HMS and HPO-MME, are used to test the transferability from knowledge-based simulated environments to real-world scenarios. Specifically, we build a \emph{synthetic training} dataset by the aforementioned process for SymCAT, utilizing the HPO entries that match the symptoms and diseases in HPO-HMS and HPO-MME, which are \emph{real-world testing} datasets. This evaluation approach can assess how well a symptom checking method generalizes when trained solely on synthetic data, because large amounts of \emph{real-world training} data are difficult to obtain.

\subsection{Experimental Protocol}

\subsubsection{Baselines}

We have selected the most appropriate RL baseline for every dataset. \textbf{GAMP} \cite{xia2020generative} uses GAN to implement an RL agent for symptom checking, which has been proven to be state-of-the-art only on tiny feature spaces (MuZhi and Dxy). \textbf{REFUEL} \cite{peng2018refuel} is a competitive RL method designed for symptom checking on a medium size search space. 

\begin{table*}
\caption{Performance of REFUEL \cite{peng2018refuel} and \mname\ on the synthetic SymCAT dataset and HPO-based rare disease datasets.}  
\label{acc}

\begin{tabular}{ccccccccccc}  
\toprule  
\multirow{2.5}{*}{Dataset}&\multirow{2.5}{*}{\shortstack{\#-Disease\\Task}}&\multirow{2.5}{*}{\#Symptoms}&
\multicolumn{4}{c}{REFUEL \cite{peng2018refuel}}&\multicolumn{4}{c}{\mname}\\
\cmidrule(lr){4-7} \cmidrule(lr){8-11}  
&&&Top1&Top3&Top5&\#Rounds&Top1&Top3&Top5&\#Rounds\\
\midrule 
\multirow{3}{*}{\shortstack{SymCAT-\\Synthetic}} & 200 & 328 & 53.76 & 73.12 & 79.53 & 8.24 & $\mathbf{55.65\pm 0.25}$ & $\mathbf{80.71\pm 0.26}$ & $\mathbf{89.32\pm 0.29}$ & $12.02\pm 0.06$\\
& 300 & 349 & 47.65 & 66.22 & 71.79 & 8.39 & $\mathbf{48.23\pm 0.25}$ & $\mathbf{73.82\pm 0.32}$ & $\mathbf{84.21\pm 0.17}$ & $13.10\pm 0.05$\\
& 400 & 355 & 43.01 & 59.65 & 68.89 & 8.92 & $\mathbf{44.63\pm 0.29}$ & $\mathbf{69.22\pm 0.11}$ & $\mathbf{79.54\pm 0.15}$ & $14.42\pm 0.03$\\
\midrule
\multirow{2}{*}{HPO-Synthetic} & 500 & 1901 & 64.33  & 73.14  & 75.34  & 8.09  & $\mathbf{76.23\pm 0.47}$ & $\mathbf{84.17\pm 0.53}$ & $\mathbf{86.99\pm 0.41}$ & $5.34\pm 0.04$\\
& 1000 & 3599 & 40.08  & 62.67  & 67.42  & 14.19  & $\mathbf{67.03\pm 0.29}$ & $\mathbf{75.94\pm 0.53}$ & $\mathbf{79.27\pm 0.58}$ & $10.62\pm 0.08$\\
\midrule
HPO-HMS & 37 & 747 & 18.22  &34.07  & 45.31 & 11.26 &  $\mathbf{34.22\pm 0.93}$ & $\mathbf{61.39\pm 1.12}$ & $\mathbf{68.43\pm 1.12}$ & $13.40\pm 0.06$\\
HPO-MME & 18 & 559 & 48.15 & 58.40 & 67.33 & 9.23 &  $\mathbf{60.70\pm 1.28}$ & $\mathbf{79.53\pm 1.40}$ & $\mathbf{83.37\pm 0.86}$ & $10.14\pm 0.14$\\
\bottomrule  
\end{tabular}  

\end{table*}

\subsubsection{Experimental Design}

\mname\ first trains the knowledge-guided self-attention model in the diagnosis branch, and then trains the VAE model with a fraction of the features dropped at random in the inquiry branch. In comparison to RL, \mname\ handles this training naturally.

For the synthetic datasets and the two rare disease datasets, HPO-HMS and HPO-MME, a positive symptom is chosen uniformly at random as a patient self-report at the beginning of the inquiring process. MuZhi and Dxy already have self-reports. For all real-world datasets, the filtering speedup technique is not applied because their numbers of candidate symptoms are already small. 
We show the results of \mname\ in percentage with a 95\% confidence interval for 5 and 20 random runs, respectively for the synthetic and real-world datasets.

\begin{figure}
  \centering
  \includegraphics[width=0.7\linewidth]{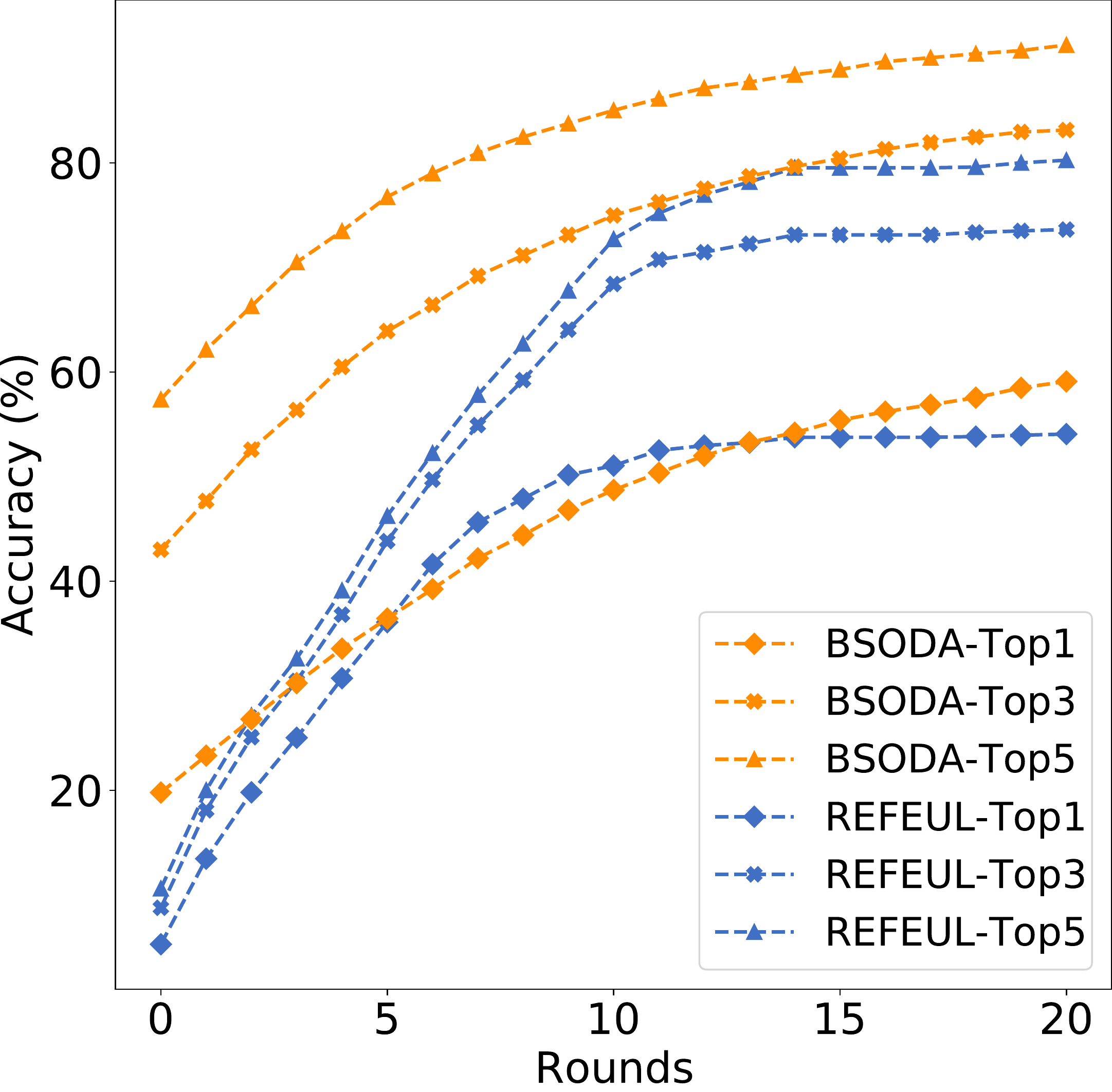}
  \caption{Accuracy vs the maximum number of inquiries for \mname\ and REFUEL \cite{peng2018refuel} on the  SymCAT 200-disease task.}
  \Description{Within the beginning several steps, the accuracy curve of REFUEL is far below the one of \mname. When the accuracy curve of REFUEL approaches its average steps (8.24), the slope of the curve is going down, and when it goes over 10, the slope is nearly zero. On the contrary, the accuracy curve of \mname\ maintains a quick growth rate. For Top3 and Top5, the curves of REFUEL are always under the ones of \mname. Although there are intersections in Top1 curves, \mname\ would outperform REFUEL at the beginning and the future.}
  \label{fig:steps_rl}
\end{figure}

\subsection{Results}

\subsubsection{SymCAT}

We synthesize three diagnostic tasks containing 200, 300, and 400 diseases, respectively. We set $N_M$ to 100 for \mname\ and the maximum number of inquiries to 15, 16, and 17 for these three tasks respectively for both \mname\ and REFUEL. The results are reported in Table~\ref{acc}. We observe that on average, \mname\ conducts more inquiries than REFUEL, and as a result, it has a slightly higher accuracy for the Top 1 prediction and a significantly higher accuracy for the Top3 and Top5 predictions.

In Figure~\ref{fig:steps_rl}, we present how accuracy changes as the maximum number of inquiries increases for the 200-disease task. REFUEL was trained to make a maximum of 20 inquiries. We chose the predicted disease with the highest action value under the learned policy. In Figure~\ref{fig:steps_rl}, we observe that the accuracy curves for REFUEL are significantly lower than those for \mname\ in the initial several inquiries. It reveals that with the partial observed information of patients, REFUEL may struggle to make accurate predictions, which is where \mname's diagnostic method that leverages the generative model to handle the uncertainties excels.

When REFUEL's accuracy curves approach their optimal number of inquiries, the curves reach a plateau. In comparison, \mname's accuracy curves continue to grow at a relatively faster rate. \mname\ always performs better than REFUEL on the Top3 and Top5 curves. Although the Top1 curves intersect, \mname\ outperforms REFUEL at the beginning and the end. REFUEL stops the inquiry and performs a prediction when it believes that it has sufficient evidence; however, \mname\ proves that this decision may not be optimal: with a few additional inquiries, \mname\ may provide a significantly better disease prediction, which is critical in the medical domain. In conclusion, using the same maximum number of inquiries, an RL agent would converge sooner and make an inferior prediction. It is mostly likely due to the difficulties to determine whether the exploration, i.e., the information gathered, is sufficient to provide a "mature" diagnosis for RL agents. In comparison, the heuristic stop criterion of \mname\ has an advantage.

\begin{table}\small
\caption{Disease prediction accuracy (\%) of REFUEL \cite{peng2018refuel}, GAMP \cite{xia2020generative} and \mname\ on two medical dialogue datasets.}  
\label{real}
\begin{center}
\begin{tabular}{cccc}  
\toprule  
 & REFUEL \cite{peng2018refuel} & GAMP \cite{xia2020generative} & \mname \\
 
\midrule  
MuZhi \cite{wei2018task} & 71.8 & 73 & $73.1\pm 0.5$\\
Dxy \cite{xu2019end} & 75.7 & 76.9 & $\mathbf{80.2\pm 0.3}$\\
\bottomrule  
\end{tabular}
\end{center}
\end{table}

\subsubsection{Real-World Common Disease Datasets}

Table~\ref{real} shows the results for REFUEL, GAMP and \mname\ on the MuZhi and Dxy datasets. For \mname, we set $N_M$ to 100 and the maximum number of inquiries to 16 for MuZhi and 20 for Dxy, respectively. For both GAMP and REFUEL, we limit the maximum number of inquiries to 20. The results in Table~\ref{real} show that among these three approaches, \mname\ has a competitive performance on MuZhi and achieves the highest accuracy on Dxy.

From the Dxy dataset, we conducted a case study to investigate the actual inquired symptoms of REFUEL, \mname\ and a doctor. The patient selected was a 21-month infant, who initially reported \emph{Coughing Up Phlegm}, and was easily misdiagnosed as having \emph{Upper Respiratory Infection}. The results are shown in Table~\ref{case} with positive inquired symptoms highlighted in bold. The key implicit symptoms were \emph{Moist Rales} and \emph{Difficult Breathing}. We observe that REFUEL provided a hasty and incorrect diagnosis without inquiring about the key symptoms. \mname\ tended to make a differential diagnosis by first inquiring \emph{Nose Rubbing} and \emph{Allergic Symptoms} in order to exclude \emph{Allergic Rhinitis}, and then inquiring about the key symptoms to provide a correct diagnosis.

\subsubsection{Proposed HPO-Based Rare Disease Datasets} We utilize the HPO to generate synthetic tasks with thousands of features to evaluate the performance of symptom checking methods. Two synthetic tasks were generated, containing 500 and 1,000 diseases, respectively. Our goal is to evaluate the models in real-world scenarios with a symptom count six times larger than that of SymCAT, the previous largest dataset. For \mname, we set the maximum number of inquiries to 10 and 18 for the tasks with 500 and 1,000 diseases, respectively, and fix $N_M$ to 20. In training REFUEL, we set the maximum number of inquiries to 14 and 18 for these two tasks, respectively. The results of HPO-synthetic in Table~\ref{acc} show that \mname\ outperforms REFUEL by a large margin, with accuracy from 11 to 27 percentage points higher while making fewer inquiries. When the number of diseases grows from 500 to 1,000, the Top1 accuracy of REFUEL drops by 24\%, but that of \mname\ decreases by only 9\%. Thus, \mname\ shows clearer advantages over RL when dealing with large feature spaces.

For the HPO-HMS and HPO-MME datasets, we set the maximum number of inquiries to 15 for both REFUEL and \mname, and fix $N_M$ to 100 for \mname. As shown in Table~\ref{acc}, \mname\ outperforms REFUEL by from 12 to 21 percentage points. However, the results on the HPO-HMS dataset are not satisfactory enough since the Top1 accuracy of \mname\ is only 34. Indeed, because of the disparity in symptom distributions between the knowledge base and real-world data, transferability of symptom checking methods remains to be a challenge.  

\begin{table}\small
\caption{A case study for actual inquired symptoms on the Dxy dataset. The patient was a 21-month infant who initially reported \emph{Coughing Up Phlegm}.}  
\label{case}

\begin{tabular}{ccc}  
\toprule  
Source & Inquired Symptoms & Diagnosis\\
\midrule  
REFUEL & Runny Nose, Green Stool, Fever, & Upper Respiratory \\
\cite{peng2018refuel} & Excessive Breathing & Infection \\
\midrule  
\multirow{3}{*}{\mname} & Fever, Nose Rubbing, & \multirow{3}{*}{Pneumonia}\\
 &  Allergic Symptoms, \textbf{Moist Rales}, & \\
 &  \textbf{Difficult Breathing}, Runny Nose  & \\
\midrule  
Doctor & \textbf{Moist Rales}, \textbf{Difficult Breathing}   & Pneumonia \\
\bottomrule  
\end{tabular}  

\end{table}

\subsection{Ablation Study of Speedup Techniques}

We performed an ablation study to show the acceleration provided by various components of the proposed speedup scheme. We ran \mname\ on the synthetic SymCAT 200-disease task, and calculated the average CPU time per inquiry with various speedup settings. We set $N_M$ to 100 and the maximum number of inquiries to 15. In Table~\ref{speed}, we present the results produced using an NVIDIA GeForce GTX 1080Ti GPU. The results demonstrate that while using the symptom filtering strategy alone achieves about 1.59x speedup, 
combining the symptom filtering and approximate sampling strategies result in a much larger speedup of 70x 
. Obviously, adopting the approximation strategy alone results in a significant speedup, and without them, the reward estimation process of EDDI \cite{ma2019eddi} is impracticable for an online service.

\section{Limitations}

From the view of \mname, a limitation is that the inquiry branch does not leverage medical knowledge explicitly. Although we use a knowledge-guided diagnosis model to assist inquiring, the model-driven symptom-inquiry process is likely to differ from the way that doctors interact with patients. This is for our future study.

From the view of symptom checking research, a limitation is that there is lack of evaluation about co-morbidity, i.e., a patient having two or more diseases, which is common in the real world. Ideally, we could simply extend \mname\ to handle co-morbidity by assuming that the diseases are independent. All that is required is to train a multi-label diagnosis model instead of a multi-class one. Unfortunately, we lack evaluation data in this area since we could not find co-morbidity datasets in current literature. Another limitation in symptom checking research is that using synthetic datasets frequently results in inflated performance. It is because the data generation models make a very simple assumption of symptom-disease and symptom-symptom relationships, while the real-world cases are much more complex.

\begin{table} 
\caption{Average CPU time per inquiry with various speedup settings for \mname\ on the SymCAT 200-disease task.}  
\label{speed}

\begin{tabular}{cc}  
\toprule  
Settings & Time (s)\\
\midrule  
\mname & $\mathbf{0.45}$\\
\mname\ w/o Approximation & 20.47\\
\mname\ w/o Approximation \& Filtering & 32.52\\
\bottomrule  
\end{tabular}  

\end{table}

\section{Conclusion}

We offer \mname, a novel non-RL bipartite framework for online disease diagnosis that is computationally efficient and scalable to large feature spaces. We adopt a Product-of-Experts encoder to efficiently handle partial observations of symptom data and design a two-step sampling strategy to estimate the reward more precisely. We develop filtering and approximation strategies to significantly speed up estimations of the information reward to a level that is practical for online services. As demonstrated by tests, \mname\ outperforms prior methods in their respective settings \cite{peng2018refuel,kachuee2018opportunistic,ma2019eddi,xia2020generative}. With a novel evaluation method, \mname\ demonstrates its ability to effectively handle large search spaces. According to our knowledge, this is the first work that proposes to test the transferability of symptom checking methods from knowledge-based simulated environments to real-world data. 

\begin{acks}
This work is supported by the National Natural Science Foundation of China (grants 61872218), National Key R\&D Program of China (2019YFB1404804 and 2021YFF1200009), Beijing National Research Center for Information Science and Technology (BNRist), Tsinghua University Initiative Scientific Research Program, and Guoqiang Institute, Tsinghua University. The funders had no roles in study design, data collection and analysis, the decision to publish, and preparation of the manuscript.
\end{acks}

\bibliographystyle{ACM-Reference-Format}
\bibliography{1}


\appendix

\section{Overview of \mname}

Please refer to Algorithm \ref{alg}.

\begin{algorithm}
\label{alg}
\caption{Overview of \mname.}  
    
\begin{algorithmic}  
\REQUIRE Training dataset $\mathbf{X}$; Test dataset $\mathbf{X}^*$ with no observations; Indices $D$ of diseases.
    
\STATE 1: \textbf{Train knowledge-guided self-attention model on complete $\mathbf{X}$ and train VAE with PoE encoder on $\mathbf{X}$ with a fraction of features dropped at random.}
\STATE 2: \textbf{Actively inquire symptom to estimate $\mathbf{x}_D$ by $p_D(\cdot)$ for each test point:}

\FOR{each test point in $\mathbf{X}^*$}
\STATE $\mathbf{x}_O\leftarrow \{x_i\}$ (initial symptom $i$)
\STATE $S_c\leftarrow S_i$
\REPEAT
\FOR{every symptom $s$ in $S_c$}
\STATE // \textit{Two-step sampling }
\STATE $\hat{\mathbf{z}}\sim q_\Phi(\mathbf{z}|\mathbf{x}_O)$
\STATE $p_{VAE}\leftarrow p_\theta(x_s|\hat{\mathbf{z}})$
\STATE $\hat{x}_s\leftarrow 1$ // \textit{Focusing on positive symptoms}
\STATE $p_{Diag}\leftarrow p_D({\hat{x}_s, \mathbf{x}_O})$
\STATE $p(\mathbf{x}_D, x_s|\mathbf{x}_O)\leftarrow p_{VAE}\cdot p_{Diag}$  (Sec.~\ref{sec:two-step})
\STATE \ 
\STATE // \textit{Approximation of the reward}
\STATE $\{\hat{R}_{j}^{k}\}_{j=0,1;k=1,...,|D|}\leftarrow \{0\}$
\FOR{disease $d$ with probability in the 90 percent of the top probability in $p_{Diag}$ and more than $1 / |D|$}
\STATE $\hat{R}_{1}^{d}\leftarrow$ Eqn. \ref{eq:r}
\ENDFOR
\STATE $\hat{R}_s\leftarrow$ a weighted average by $p(\mathbf{x}_D, x_s|\mathbf{x}_O)$ and $\{\hat{R}_{j}^{k}\}$
\ENDFOR
\STATE $t\leftarrow\text{argmax}_t\hat{R}_t$, \textbf{inquire symptom $t$}
\STATE $\mathbf{x}_O\leftarrow \mathbf{x}_t\cup \mathbf{x}_O$
\IF{$\mathbf{x}_t$ is positive}
\STATE $S_c\leftarrow S_t\cap S_c$ // \textit{Filtering candidate symptoms}
\ENDIF
\STATE $\hat{\mathbf{z}}\sim q_\Phi(\mathbf{z}|\mathbf{x}_O), \hat{\mathbf{x}}_{U}\sim p_\theta(\mathbf{x}_U|\hat{\mathbf{z}})$
\STATE $\mu_{p_D} \leftarrow \mathbb{E} [p_D(\mathbf{x}_O\cup \hat{\mathbf{x}}_{U})], \sigma_{p_D} \leftarrow \sqrt{\mathrm{Var}[p_D(\mathbf{x}_O\cup \hat{\mathbf{x}}_{U})]}$
\STATE $d\leftarrow\text{argmax}_d\ \mu_{p_D}^{(d)}$, \textbf{report disease $d$ }
\UNTIL{reaching maximum number of inquiries $N_T$ or $\forall j\neq d, \mu_{p_D}^{(d)}>\mu_{p_D}^{(j)}+3\sigma_{p_D}^{(j)}$}
\ENDFOR

\end{algorithmic}  
\label{alg}
\end{algorithm}

\section{Study with Hybrid Data Types}

\citet{kachuee2018opportunistic} compiled a diabetes dataset from the national health and nutrition examination survey (NAHNES) data \cite{nhr}. It consists of 45 features, such as demographic information, lab results (total cholesterol, triglyceride, etc.), examination data (weight, height, etc.), and questionnaire answers (smoking, alcohol, etc.). In contrast to other symptom checking tasks, these features are continuous, and each feature is associated with a cost suggested by an expert in medical studies. Finally, the fasting glucose values were used to define three classes based on standard threshold values: normal, pre-diabetes, and diabetes. The dataset consists of 92,062 instances and is randomly split into three subsets, including 15\% for testing, 15\% for validation, and the rest 70\% for training.

\begin{figure}
  \centering
  \includegraphics[width=0.9\linewidth]{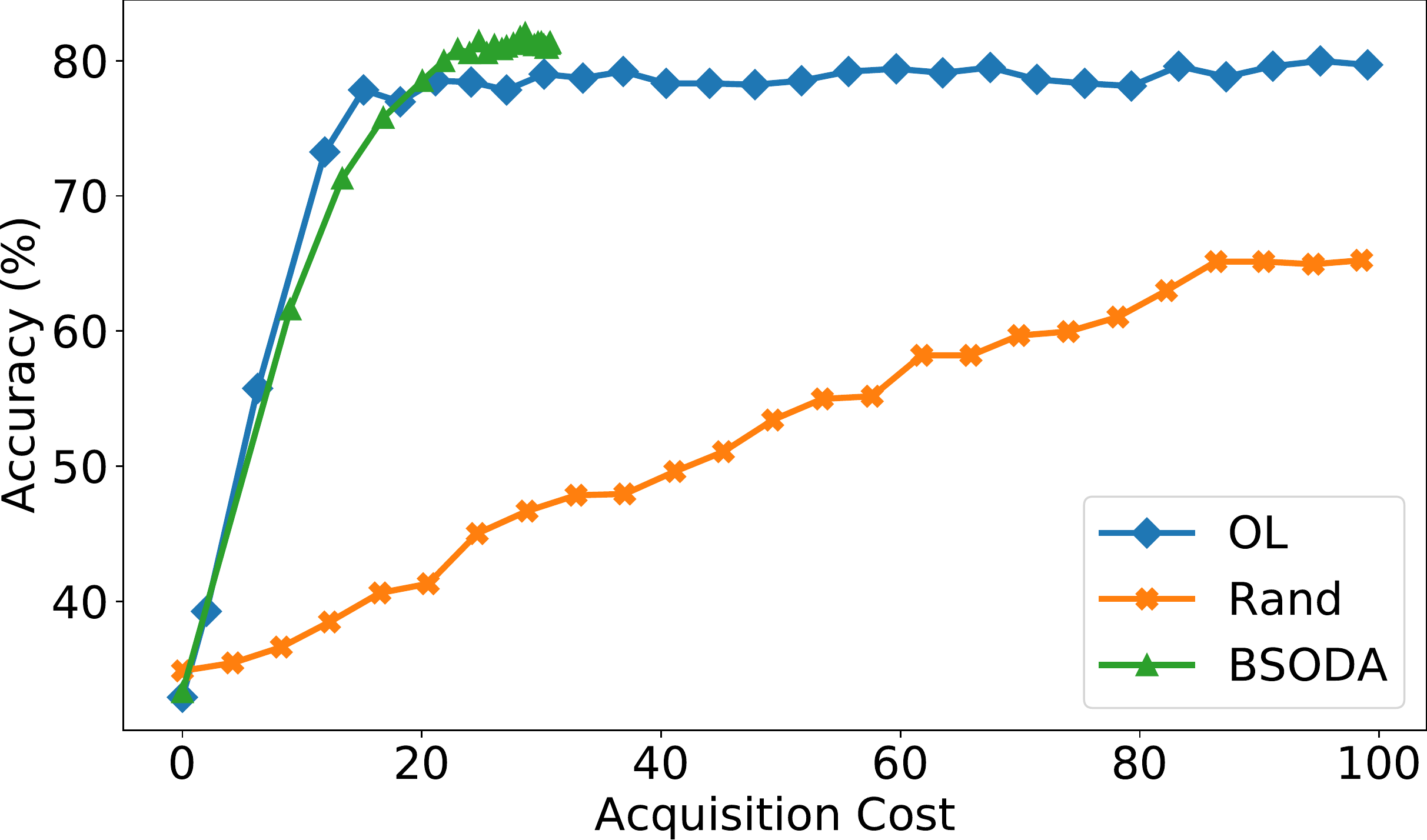}
  \caption{Accuracy vs acquisition cost for \mname, OL \cite{kachuee2018opportunistic} and a random method on the diabetes dataset.}
  \Description{\mname\ achieves a superior accuracy with a lower acquisition cost than OL.}
  \label{fig:db}
\end{figure}

Figure \ref{fig:db} shows an evaluation on the diabetes dataset that has hybrid types of features. We implemented \mname\ without using any speedup techniques to handle these continuous features. We set $N_M$ to 50 for \mname\ and did not restrict the maximum number of inquiries. When the maximum number of inquiries is set to $i$, the $i$-th point on the curve plots the average cost (x-axis) and the accuracy (y-axis). We observe that \mname\ achieves a superior accuracy while having a lower acquisition cost than OL \cite{kachuee2018opportunistic}, which is an RL baseline based on deep Q-learning and designed for cost-sensitive feature selection. Except for the diabetes dataset, OL is not optimal for the other symptom checking datasets because their feature costs are unknown. Thus, in addition to symptom checking, \mname\ has the ability to handle online diagnosis problems involving hybrid data types.

\begin{figure*}
  \centering
  
  \includegraphics[width=\linewidth]{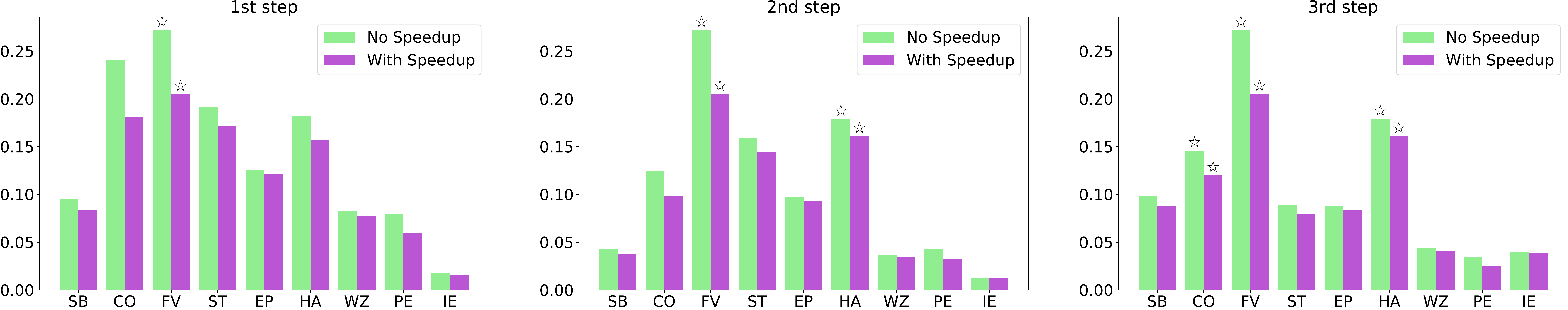} 
  \caption{Partial distributions for information rewards among several symptoms in first three inquiries on the SymCAT 200-disease task. The symptoms with maximum reward are marked with star. The initial symptom is "Nasal Congestion". "SB" means "Shortness of Breath". "CO" means "Cough". "FV" means "Fever". "ST" means "Sore Throat". "EP" means "Ear Pain". "HA" means "Headache". "WZ" means "Wheezing". "PE" means "Pain in Eye". "IE" means "Itchiness of Eye".}
  \label{fig:dis}
\end{figure*}

\section{Case Study for Speedup Techniques}

We conducted a case study to show that the proposed speedup techniques have little effect on the information reward and the actual symptoms queried by \mname. Figure \ref{fig:dis} shows the partial distributions of information rewards for several symptoms in the first three inquiries for a case. The test case comes from the synthetic SymCAT 200-disease task and is a simulated patient with \emph{Common Cold}. We limit the maximum number of inquiries to 15 and $N_M$ to 100.

\begin{table*}
\caption{Accuracy (\%) with various model settings for \mname\ on the SymCAT 200-disease task.}  
\label{ab}

\begin{tabular}{ccccc}  
\toprule  
Settings & Top1 & Top3 & Top5 & \#Rounds \\
\midrule
\mname & $\mathbf{55.65\pm 0.25}$ & $\mathbf{80.71\pm 0.26}$ & $\mathbf{89.32\pm 0.29}$ & $12.02\pm 0.06$\\
\mname\ w/o Diag. & $47.37\pm 0.50$ & $72.28\pm 0.57$ & $82.02\pm 0.81$ & $14.31\pm 0.03$ \\
\mname\ w/o Diag. \& Inq. & $30.36\pm 0.26$ & $56.42\pm 0.36$ & $70.70\pm 0.36$ & $4.21\pm 0.10$ \\
\bottomrule  
\end{tabular}  

\end{table*}

\section{Ablation Study of Models}

We performed an ablation study to investigate the effect of model design on the synthetic SymCAT 200-disease task. We set $N_M$ to 100 and the maximum number of inquiries to 15. In Table~\ref{ab}, "Inq." and "Diag." refer to the symptom-inquiry and disease-diagnosis model designs in \mname, respectively. "\mname\ w/o Diag." means we replace the predictive distributions $p_D(\cdot)$ from the diagnosis model in the two-step sampling strategy and the diagnostic method with ones produced by the VAE model. "\mname\ w/o Inq." means we do not use the PoE encoder. "\mname\ w/o Diag. \& Inq." is simply the EDDI framework. The results show that the PoE encoder improves the accuracy significantly by at least 12\% and the average number of inquiries increases. With the same termination criterion and the maximum number of inquiries, the PoE encoder provides a better and more patient model, which does not diagnose hastily. Besides, the diagnosis model is critical in effective diagnosis with fewer inquiries and providing a narrower confidence interval, indicating a more stable result.

\end{document}